# A Novel Machine Learning Method for Preference Identification

**Azlan Iqbal** AZLAN@UNITEN.EDU.MY
*College of Computing and Informatics, Universiti Tenaga Nasional*
*Putrajaya Campus, 43000 Kajang, Selangor, Malaysia*

**Abstract**

Human preference or taste within any domain is usually a difficult thing to identify or predict with high probability. In the domain of chess problem composition, the same is true. Traditional machine learning approaches tend to focus on the ability of computers to process massive amounts of data and continuously adjust 'weights' within an artificial neural network to better distinguish between say, two groups of objects. Contrasted with chess compositions, there is no clear distinction between what constitutes one and what does not; even less so between a good one and a poor one. We propose a computational method that is able to learn from existing databases of 'liked' and 'disliked' compositions such that a new and unseen collection can be sorted with increased probability of matching a solver's preferences. The method uses a simple 'change factor' relating to the Forsyth-Edwards Notation (FEN) of each composition's starting position, coupled with repeated statistical analysis of sample pairs from both databases. Tested using the author's own collections of computer-generated chess problems, the experimental results showed that the method was able to sort a new and unseen collection of compositions such that, on average, over 70% of the preferred compositions were in the top half of the collection. This saves significant time and energy on the part of solvers as they are likely to find more of what they like sooner. The method may even be applicable to other domains such as image processing because it does not rely on any chess-specific rules but rather just a sufficient and quantifiable 'change' in representation from one object to the next.

## 1. Introduction

Machine learning, a term popularized by Arthur Lee Samuel (Samuel, 1959), essentially encompasses computational approaches that help us find meaningful patterns in data; especially without the need to explicitly program the computer to do so. It is generally preferable where traditional rule-based programming does not work as well or at all. Perhaps one of the most dramatic contemporary examples of machine learning progress was the 2016 five-game match between Lee Sedol, perhaps the top go player at the time, and AlphaGo, a computer program designed by Google's 'DeepMind' company (Metz, 2016). AlphaGo defeated Lee Sedol four games to one; something some experts believed would not happen for another decade or so (Associated Press, 2016; Yan, 2016). Soon after that, their 'AlphaZero' program was generalized enough to be able to *master* the games of chess, shogi and go (Silver et. al, 2018).

In chess, the then world champion, Garry Kasparov, was defeated about two decades earlier in a 1997 match by IBM's 'Deep Blue' computer (Hsu, 2002). It did not use machine learning like 'AlphaZero' but a brute-force, heuristics-based approach. Since then, chess programs have served more as tools for humans, even master players, to train with rather than compete against (Dhou, 2008; Khadilkar, 2019). A chess app downloaded to a smartphone today is stronger than most grandmasters, and if running on a typical desktop or even notebook computer could likely defeat even the world champion (Newborn, 2011). Unfortunately, this has led some master players to cheat (Friedel, 2017; Chiu, 2019). The subdomain of chess composition, however, remains relatively immune to AI or machine learning advances.

We are not referring to the *solving* of such problems (existing chess engines do that quite well, if not perfectly, already) but rather the assessment of human preference with regard to aesthetics, interestingness and qualities of that nature. Aspects of chess compositions that are difficult, if not impossible, to quantify but that appeal to people on an individual basis regardless of domain knowledge and experience. In short, is there a way for a computer to learn from a collection of one's personal liked and disliked compositions such that *new* compositions can be at least sorted in terms of the probability that they will also be liked by the solver? In the following section, we briefly review the various machine learning categories and 'recommendation system' approaches. In section 3, our methodology is explained in detail. In section 4, we present our experimental results. Finally, we conclude the article in section 5 with a discussion of the results and some suggestions for future work.

## 2. Review

In general, there are four main categories of machine learning and two main approaches used in recommender systems that we would like to briefly explain here. (In this article, the terms chess problem, chess puzzle and chess composition are used interchangeably; 'user' and 'solver' refer to the subject or person in question.)

### 2.1 Machine Learning Categories

The existing categorizations of machine learning, i.e. supervised, unsupervised, transfer and reinforcement were deemed unsuitable for preference identification in chess compositions. Supervised learning relies on proper labeling of many examples. For instance, in distinguishing between photos of cats and dogs. Chess compositions cannot be reliably labeled in this way. While they can be easily categorized as valid or invalid – a composition is valid if its stipulation (e.g. 'White to play and mate in 3 moves') is found to be true based on say, exhaustive chess software engine analysis – that is just the tip of the iceberg. A particular chess problem may contain one or more chess themes (e.g. pin, fork, skewer) and abide by one or more composition conventions (e.g. no 'check' in the first move, no duals); it could also be paradoxical which means it goes against what is typically taught to chess players (Howard, 1967; Levitt and Friedgood, 2008). Not to mention possessing other qualities such as economical use of pieces and being challenging but not too complex (Margulies, 1977). So, it is not 'one thing' as opposed to 'another thing' such as a cat versus a dog.

Unsupervised learning attempts to find patterns in data without labels and is typically used in anomaly detection. For instance, likely fraudulent credit card transactions (Rai and Dwivedi, 2020). Anomalies or paradoxical moves in chess problems (and even regular games), on the other hand, are typically sought after, e.g. an unexpected queen or major piece sacrifice (Shenk, 2006). Compared to the original position, a single piece moved to a neighboring square may not look very different or 'anomalous' to an unsupervised learning system but it could change everything about a chess problem or completely invalidate the solution. Reinforcement learning uses 'rewards' and 'penalties' in order for the system to learn the rules or 'policies' which can be effective in learning to *play* chess (like with 'AlphaZero') but not really to evaluate what makes a good chess problem or puzzle, especially from the viewpoint of a particular person. This is because the rules of playing and what constitutes a win in a game are usually quite clear whereas what qualifies in terms of personal taste is not. For example, a solver could specify any number of conventions that a chess problem must satisfy yet still encounter many that are not appealing, and furthermore actually miss many that do not satisfy all those conventions yet would have been appealing for other reasons that may not even be explicable by the solver.

Transfer learning relies on using the rules from another system and applying it to your own (Zhuang et al., 2020). In practice, this may involve using the existing and already-trained lower layers of a neural network in some form of say, image detection and simply adding another higher (prediction) layer which helps describe a higher-level image. The new system 'reuses' the lower layers already trained. For chess problems, this would mean depending on an existing simpler and related system that can be applied. Some people might say that 'chess puzzles' are not as sophisticated as 'chess problems' or that the conventions are not as many, even though their definitions imply they are generally interchangeable or at least overlap significantly (Iqbal, 2019). So, if we had an existing set of neural network layers describing chess puzzles, we could simply transfer them to a more sophisticated chess *problem* network with an additional layer.

Unfortunately, there is no known trained system for chess puzzles to transfer learning from in this case. Furthermore, distinguishing between chess problems and chess puzzles would be difficult, if not impossible, since many 'simpler' or less sophisticated chess problems would closely resemble chess puzzles. A chess puzzle composer may condone or accept having a 'check' in the first or key move, for instance, yet there are also renowned chess problems where the key move starts with a 'check' (Friedel, 2018). This is because exceptions are not uncommon in problem composition, especially if there are compensating factors. At this point, we have not even factored in personal taste or preference yet. Therefore, even transfer learning is unsuitable here.

## 2.2 Recommendation System Approaches

Recommender or recommendation systems more directly address the aspect of user preference or taste in a variety of domains (Deldjoo et al., 2020). Two main difficulties arise. The first is that new objects to be evaluated are unseen and unpredictable (i.e. they arise in the future) and the second is that a person's taste or preferences can, and probably will, change with time. One approach is to rely on content filtering or asking users what they like and dislike. For instance, with regard to films, a user may indicate that they like drama and action movies but dislike comedies and musicals. Therefore, a new film which has been classified as largely a mix of action, some drama and a little comedy may be recommend to this user. A new film which is mostly a musical comedy with some action, on the other hand, will not be recommended.

Gathering all this feature data can be cumbersome. It is also problematic because the system only improves (to a point) given more and more features including further details about each user and their preferences, e.g. user nationality, level of education, favorite actors, preferred movie length. This can be burdensome to the user and they are not always particularly good or accurate at describing their preferences either. Often, a user does not know if they will like something until they actually experience it and even then, they are not always able to explain why in a way that would make sense to someone collecting or processing the data.

Another approach is to filter 'collaboratively', i.e. to rely on unknown or 'latent features' that arise from the data (Khenissi et al., 2020). This may refer to the data about users and what they actually watch (and for how long etc.). In practice, because many other 'similar users' also watched another movie, you might find the system recommending it to you as well. The system is able to learn from the patterns found in the data because the data are not random. It is essentially a more precise, effective and dynamic approach to the problem. The main issue with existing recommendation approaches like these with regard to chess compositions is that there is no known service where a sufficiently large user base experiences them in the way they experience say, music or movies. Even if there is (e.g. an online chess community featuring chess puzzles that can be rated), it is not a localized and personalized learning system that each user can run on their own preference data. Data that are not intermingled or 'contaminated', to an extent, with the preferences of other 'similar people'.

## 3. Methodology

In this section, we explain progressively and in detail the various aspects of our proposed novel machine learning method.

### 3.1 Detecting Changes Between Objects

The initial position of a chess problem (the 'object', in this case) is typically recorded using Forsyth-Edwards Notation (FEN). Figure 1 shows two example positions with their corresponding FENs.

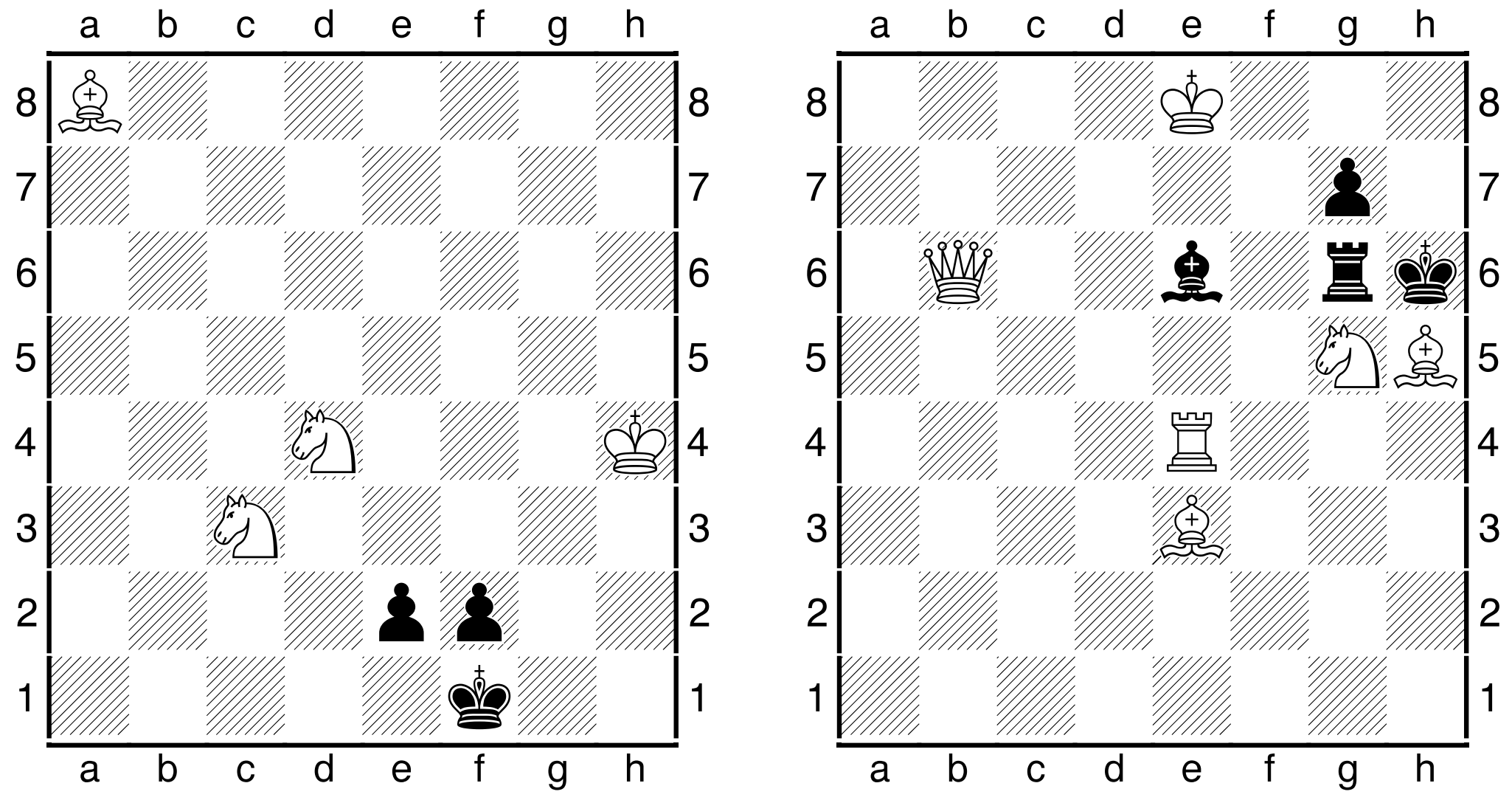

B7/8/8/8/3N3K/2N5/4pp2/5k2 w - - 0 1
(a)

4K3/6p1/1Q2b1rk/6NB/4R3/4B3/8/8 w - - 0 1
(b)

Figure 1: Example chess positions with their FENs.

Each row, starting from the upper left corner is separated by a slash ('/'). Within each row, uppercase letters are used for the white pieces and lowercase letters for the black pieces, e.g. 'K', 'q', 'R', 'b',

‘N’, ‘p’. Empty squares are represented using a digit, such as ‘1’ for a single empty square between two pieces or ‘3’ for a stretch of three empty squares. This is followed by a space and either a ‘w’ or ‘b’ to indicate the side to move. The remaining characters have to do with castling permissions, possible ‘en passant’ capturing, and the number of half-moves (plies) and full moves played. More information about this format can be found in (Wikipedia, 2020). Everything starting with the side to move is irrelevant for our purposes in this article.

In order to detect changes between FENs, they need to be converted, up to the point of the bottom row information, into a form that is more consistent in length. This is achieved by simply converting all the digits that represent empty squares into ‘1s’ and removing the slashes as well because they cannot change from one position to the next. So, the FEN for Figure 1(a), for instance, would now look like this (but without even the blank spaces every eight characters which have been added here for legibility): ‘B1111111 11111111 11111111 11111111 111N111K 11N11111 1111pp11 11111k11’. The same process applied to the FEN for Figure 2(b) would ensure a string of the same length where each character can be compared against the one in Figure 1(b) or any other FEN consistently. This form sufficiently represents the ‘look’ of the position.

The change from one FEN to the next is based on two simple rules. The first is that if a blank character, i.e. a ‘1’ changes to any piece (e.g. ‘K’, ‘r’) or vice versa, then the ‘change factor’ (CF) as a percentage value is 100, i.e. a total change. The second rule is that if the change is from one piece to another piece (e.g. ‘K’ to ‘Q’), the CF is 50 or a partial change. A blank square to a blank square or a piece to the same piece does not count. These values are accumulated for the entire string and then summed as the ‘change value’ (CV) between the two FENs. It is noteworthy here that this technique of detecting changes does not rely on any chess-specific rules, which was our intention. It should make the process more generic and widely applicable. Table 1 shows an example of the change value from FEN 1 to FEN 2.

| | |
|---|---|
| **FEN 1** | B1111111111111111111111111111111111N111K11N111111111pp1111111k11 |
| **FEN 2** | 1111K111111111p11Q11b1rk111111NB1111R1111111B1111111111111111111 |
| **CV** | [(17/64) x 100] + [(0/64) x 50] = 26.5625 |

Table 1: Calculating the ‘change value’ (CV) from one FEN to the next.

The first and second FENs have been ‘expanded’ to their 64-character forms. Each character is compared with its counterpart. There are 17 changes out of 64 that represent a total or 100% change and none which represent a partial change. This works out to 26.5625 or 26.562 if rounded to three decimal places. (Incidentally, ‘5’ in this case being the last available digit in the series means the ‘2’ just before it stays the same and is not incremented to ‘3’ since we do not know if the ‘5’, which is precisely in the middle of ‘0’ and ‘10’ tends upward.) Using a smaller segment as another example, ‘1K1Q1’ changing to ‘1K1R1’ or ‘1K1n1’ would have a CV of (1/5) x 50 = 10. Another notable quality of this technique is that the FENs can even be reversed and the CV would be the same, i.e. it works in both directions.

### 3.2 Processing Change Values in a Sample

Between two positions or FENs, the change value (CV) as explained in the previous section is fairly easy to calculate. However, given a collection or database of chess compositions, it is necessary for the computer to be able to learn the changes across many positions, perhaps even with newer ones being added from time to time. Using a theoretical example of a sample of seven FENs, the CVs might appear as shown in Table 2.

| FEN | CV |
|---|---|
| 1 | 0 |
| 2 | 26.562 |
| 3 | 27.344 |
| 4 | 25.000 |
| 5 | 17.969 |
| 6 | 11.719 |
| 7 | 14.062 |

Table 2: The ‘change values’ in a sample of positions.

The CV from the first FEN to the second is 26.562, the CV from the second to the third is 27.344 and so forth. Since a CV can only be calculated when a FEN is compared to a prior one, the first FEN always has an associated default CV of ‘0’ (as if comparing to nothing or itself). The ordering of these positions, in a sample, is also relevant now. If a different FEN which might otherwise have appeared much later got inserted somewhere in the middle of this sample replacing an existing one, then the CVs would for this sample would no longer be the same. This is important because the method proposed uses many t-tests between many such samples. While the number of decimal places that should be used for the CVs is not prescriptive, even higher precision than what is seen here may be preferable in some cases.

### 3.3 Applying the T-test

In order to learn what a particular person’s preferences are with regard to chess problems, it is necessary to have a collection of problems that person had seen and selected, i.e. the ‘liked’ database (LD) to contrast against the problems that person had seen and rejected, i.e. the ‘disliked’ database (DD). A sample from either database would typically be larger but look like the one shown in Table 2, i.e. FENs with a ‘change value’ (CV) beside each one. A t-test can then be used to determine if the mean CV is different between the two samples (one from each database) to a statistically significant degree. In our case, the two-tailed, two-sample t-test assuming unequal variances (TTUV) at the 5% level was deemed suitable. Ideally, the same sample size should be used for both databases.

After some testing, we determined that a variable but incremental sample size of between a minimum of 30 and a maximum of 60 (chosen randomly), repeated for three cycles worked best, i.e. provided the most consistent results. More specifically, many t-tests are actually performed and this depends on the sizes of the LD and DD. For example, the LD might have a total of 200 compositions whereas the DD might have 425 compositions. In the first cycle, let us assume the random sample size is say, 32. This means the first 32 CVs of the LD are compared against the first 32 CVs of the DD using the TTUV. Figure 2 shows what these samples could look like. The CVs in it are shown for illustrative purposes only and were not derived from actual FENs. The three dots represent the missing intermediate FENs and CVs.

| FEN | CV |
|---|---|
| 1 | 0 |
| 2 | 16.562 |
| 3 | 26.344 |
| 4 | 23.000 |
| 5 | 17.969 |
| … | … |
| 32 | 15.062 |

‘Liked’ Sample

| FEN | CV |
|---|---|
| 1 | 0 |
| 2 | 27.562 |
| 3 | 17.344 |
| 4 | 26.000 |
| 5 | 19.969 |
| … | … |
| 32 | 17.162 |

‘Disliked’ Sample

Figure 2: Example of samples from two databases.

The TTUV result of the 32 CVs from the LD sample compared against the 32 CVs from the DD sample is recorded as being significant or insignificant. This is the first test (T1). The same t-test is then performed but this time a new FEN under consideration is inserted into the last slot, i.e. 32 of the ‘liked’ sample, replacing whatever FEN was there and creating a new CV using the 31st FEN and itself. This is intended to aid in determining if the new FEN ‘belongs’ in the LD. The significance of the TTUV result returned this second time (T2) is then recorded.

A change from insignificant (in T1) to significant (in T2) counts as a ‘positive’ (POS) whereas the opposite counts as a ‘negative’ (NEG). The process is then repeated but against the next sample ‘chunk’ of the DD. This means the first 32 CVs from the LD are now compared against CVs 33 to 64 of the DD with the POS or NEG values potentially increasing. The same new FEN is used for the second t-test. Figure 3 shows the general pattern of testing, with each arrow representing a TTUV between each pairing of samples.

| FEN | CV |
|---|---|
| 1 | 0 |
| 2 | 16.562 |
| 3 | 26.344 |
| … | … |
| 32 (T1)<br>32 (T2) | 15.062 (T1)<br>25.625 (T2) |
| 33 | 18.912 |
| 34 | 15.654 |
| … | … |
| 200 | 35.655 |

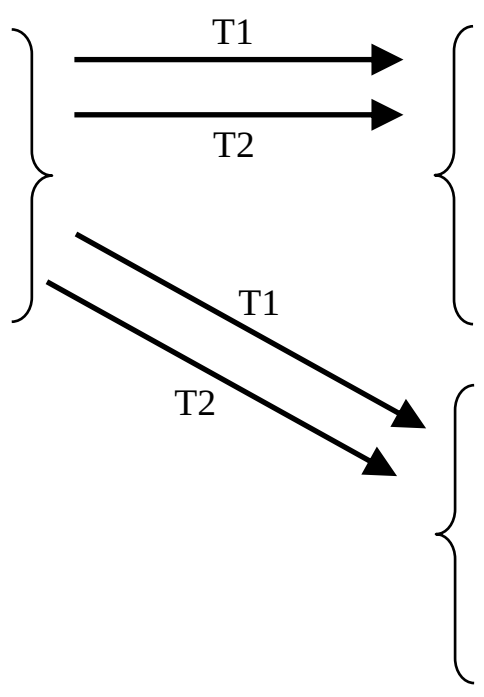

| FEN | CV |
|---|---|
| 1 | 0 |
| 2 | 27.562 |
| 3 | 17.344 |
| … | … |
| 32 | 19.969 |
| 33 | 15.424 |
| 34 | 23.000 |
| 35 | 24.265 |
| … | … |
| 64 | 17.000 |
| 65 | 15.225 |
| 66 | 17.775 |
| … | … |
| 425 | 17.162 |

Figure 3: The t-tests between samples in the databases.

In the case of this example, the testing stops at the 416th CV of the DD since another chunk of 32 is not possible given its size. So, the first 32 CVs from the LD were tested 13 times against the DD. Testing then continues in the same way as explained but with the LD sample shifting to CVs 33 to 64. This second sample chunk of the LD is compared with all the sample chunks of the DD from the first one. The sample chunks for LD stop at the 192nd CV since another chunk of 32 is not possible given its size. The total number of TTUVs (for the first cycle) is therefore 6 x 2 x 13 = 156. There will be a total of positive counts (POS) and negative counts (NEG) after all the statistical tests of all the samples from the two databases are completed in this first cycle. The ratio of the (POS) over the total counts, i.e. (POS + NEG) is returned as a percentage.

So, if POS = 36 and NEG = 14, the value obtained is [36 / (36 + 14)] x 100 = 72%. This would be the 'ranking percentage' (RP) for the new FEN under consideration in this first cycle. The entire process is repeated for two more cycles with POS and NEG reset. The second cycle, however, must have a random sample size larger than the first but never exceeding 60. One method of ensuring this over three cycles is to have a random sample size anywhere between 30 and 40 for the first cycle and for the remaining two cycles, an addition of a random number between 1 and 10 to determine the size of the new sample. For example, if '32' was selected as the first cycle's sample size, and then '5' was the random number to be added, the sample size for the second cycle would be, 32 + 5 = 37. For the third cycle, if '9' was selected at random, the third cycle's sample size for the databases would be, 37 + 9 = 45.

This incremental sample size pushes the learning toward a more stable statistical result without becoming too large (even though larger samples do speed up the process). In the end, the average rank percentage (ARP) is used. This could mean, for instance, (72% + 64% + 50%) / 3 = 62%. This value of 62% would be attributed to that single new FEN under consideration. One that is perhaps a part of a larger collection that the solver has not seen yet. All of this is no doubt time consuming computationally as hundreds, if not thousands, of TTUVs are run in order to attribute a value to a single FEN so it can be ranked. On the other hand, if there was only one new FEN that needed to be decided upon, its ARP (e.g. 62%) would be on the favorable side since it is above 50%. This means the solver might want to take a look at it. Had it been equal to or below 50%, the solver might just skip it.

### 3.4 Ranking the New Compositions

As explained in the previous section, the 'average rank percentage' (ARP) can be obtained for every new composition (with a FEN to represent its starting position). Assuming 10, 20 or even 100 new compositions are tested, they can all be ranked in descending order, i.e. with the highest ARPs naturally toward the top of the list. The idea is that *more* of what the solver is probably going to like will appear toward the top of such a list than we would expect from random chance. This kind of ranking matters because it is virtually impossible to predict with certainty which of the new compositions a particular solver is going to like. Even they cannot say for certain until seeing them.

Preference or taste also tends to change with time so a solver might suddenly come across something of a type they would not have liked before or decided now to like something of a type they perhaps did not appreciate as much before. Therefore, anything that increases the chances of the solver encountering a composition they will like in a given period of time can be helpful. This is precisely what a ranked list of compositions produced using the proposed machine learning method accomplishes. Table 3 shows what such a list might look like.

| New Composition FEN | ARP |
|---|---|
| 8/8/2Q5/1b6/1r6/5B2/k1N5/2K5 w - - 0 1 | 99.21 |
| 8/5K1k/8/8/7N/1p6/8/B7 w - - 0 1 | 97.56 |
| 3K4/6Rr/8/5B2/2Q5/8/8/3bk3 w - - 0 1 | 96.90 |
| 6Q1/8/8/1R5B/8/6R1/1Pp4K/1k6 w - - 0 1 | 93.89 |
| 1r6/8/6P1/1p1R4/k4N2/3p4/P5K1/2Q5 w - - 0 1 | 85.96 |
| 8/5R2/1k5K/2N2NR1/8/8/1p6/8 w - - 0 1 | 81.15 |
| 6R1/8/2K5/k7/8/3p4/1P6/8 w - - 0 1 | 73.61 |

Table 3: Example list of new compositions with their average rank percentages.

The particular FENs in Table 3 are not actually ranked this way but used here just to illustrate that a set of new composition FENs will have average rank percentages associated with them. Given a collection of perhaps 30 of these (only seven are shown in Table 3), if the solver would genuinely have liked only 10 of them (without any knowledge of their rankings), it is expected 7 or 8 would typically be found in the top 15 of the 30 new compositions ranked, when only 5 would be expected under normal circumstances (e.g. random selection). This is actually a 40-60% improvement. It means that in going through such a collection of ranked compositions, the solver is more likely to find significantly more of what they will like and sooner. The solver's time and energy are therefore saved given that only the top half of the collection would need to be viewed in order to find most of what they would have ended up liking.

### 3.5 The Proposed Method Algorithmically

The proposed method can be described algorithmically as follows. This is done for each new FEN for which a determination needs to be made if it is likely to be 'liked' or 'disliked' by the solver.

```
For a = 1 to 3 (i.e. cycles)
    Choose a sample size between 30 and 40 randomly (i.e. chunk size)
    For b = first chunk of 'liked' DB to last chunk of 'liked' DB
        For c = first chunk of 'disliked' DB to last chunk of 'disliked' DB
            For d = 1 to 2
                If d = 2 then
                    Replace last FEN in 'liked' sample with new FEN
                    Calculate and update the last 'change value' (CV)
                End If
                Compare 'liked' sample against 'disliked' sample using t-test
```

```
                Store t-test result
                If d = 2 then
                    Compare t-test result (of d = 1) against result (of d = 2)
                    If insignificant to significant then POS = POS + 1
                    If significant to insignificant then NEG = NEG + 1
                    Reset last FEN and CV in 'liked' sample to their original values
                End If
            Next d
        Next c
    Next b
    Store rank percentage, i.e. [(POS) / (POS + NEG)] x 100
    Add random number between 1 and 10 to sample size
    POS = 0, NEG = 0
Next a
Return average rank percentage
```

## 4. Experimental Results

We used collections of computer-generated chess problems by *Chesthetica*, an automatic chess problem composer (Chesthetica, 2020), divided into two PGN (Portable Game Notation) databases, i.e. 'liked' and 'disliked'. A PGN database is a standard format that allows chess positions and move sequences to be recorded. The 'liked' database (LD) consisted of 3,041 compositions generated between 20th July 2010 and 4th September 2020 that had been seen and *selected* by the author. The 'disliked' database (DD) consisted of 10,028 compositions generated between 20th March 2020 and 4th September 2020 that had been seen and *rejected* by the author. This does not mean that the DD contained invalid compositions but rather than the author simply did not have a personal preference or particular liking for them. Quite often, other people in chess-playing and composing communities worldwide have been known to like compositions the author does not and vice versa.

Also, there were many low-quality compositions (in the author's opinion) intentionally generated as byproducts of unrelated experimental work which is why the DD used here happens to be so much larger within a shorter period. The databases were the largest we could obtain based on availability at the time. The most recent 20 compositions from the LD were used as the 'new and unseen' sample (NUS). These had actually already been seen and preferred by the author prior to any research work on this topic but they could serve as a good representation of a new and unseen collection for testing purposes. The LD and DD are sorted based on the dates and times the compositions were automatically generated (starting with the earliest).

The aforementioned 20 happened to start from 15th August 2020, 8:47 pm. The remaining 3,021 compositions, with the last composition generated just prior to that, would therefore be used to learn from using our proposed method. The DD was truncated at composition 9,331 because everything *after* that was generated on 15th August 2020 or later. This truncated database would also be used in the learning process. So, for both the LD and DD, only subsets of compositions generated prior to all in the NUS would be used for learning. From the remaining 696 in the DD (i.e. 9,332 to 10,028), 60 were selected at random and then divided sequentially and equally into three samples of 20 as 'baseline rejected' samples (in contrast to the NUS). We labeled them as BRS A, B and C. These represent analogously 'unseen' compositions but actually already seen and *rejected* by the author prior to any research work on this topic.

An algorithmic random number generator (internal to the Microsoft Visual Basic 6 programming language) was used for all random number generation purposes. Note that random selection is not a necessity here given that all of the remaining 696 are from the DD, i.e. 60 compositions could have just as well been selected sequentially from any point early enough there. The first hypothesis is that using the proposed novel machine learning method, *more* of the NUS would be selected for computationally compared to the three BRS ones, based only on what was learned from the compositions generated prior to them. The second is that a statistically significant difference would exist between the mean 'average rank percentage' (ARP) of the NUS compared to those in each of the three BRS ones, with the mean ARP of the NUS always being higher. Table 4 shows the rank percentage (RP) score for each cycle for each FEN or position in the NUS, including the average RP.

| FEN | Cycle 1 | Cycle 2 | Cycle 3 | Avg. |
|---|---|---|---|---|
| 1 | 90.20 | 80.91 | 88.71 | 86.60 |
| 2 | 63.33 | 50.96 | 71.01 | 61.77 |
| 3 | 75.63 | 70.93 | 78.72 | 75.09 |
| 4 | 80.98 | 83.50 | 83.56 | 82.68 |
| 5 | 14.17 | 16.04 | 17.11 | 15.77 |
| 6 | 97.57 | 96.83 | 95.29 | 96.56 |
| 7 | 60.43 | 78.26 | 60.61 | 66.43 |
| 8 | 52.50 | 62.89 | 47.22 | 54.20 |
| 9 | 0.45 | 6.08 | 6.90 | 4.48 |
| 10 | 76.88 | 71.91 | 67.86 | 72.22 |
| 11 | 38.97 | 40.00 | 39.44 | 39.47 |
| 12 | 68.71 | 78.05 | 65.08 | 70.61 |
| 13 | 61.81 | 58.33 | 50.00 | 56.71 |
| 14 | 44.44 | 51.81 | 55.22 | 50.49 |
| 15 | 31.62 | 46.59 | 26.56 | 34.93 |
| 16 | 95.37 | 99.21 | 92.47 | 95.69 |
| 17 | 97.67 | 94.12 | 94.44 | 95.41 |
| 18 | 4.23 | 12.34 | 14.29 | 10.29 |
| 19 | 73.04 | 86.81 | 70.31 | 76.72 |
| 20 | 51.39 | 58.33 | 55.56 | 55.09 |

Table 4: The rank percentage scores for the 'new and unseen' sample (and the averages).

The highest, median and mean ARP values for the NUS were 96.56%, 64.1% and 60.06%, respectively. Analogous data for the BRS ones, but showing just the average rank percentages, are in Table 5.

| FEN | BRS A | BRS B | BRS C |
|---|---|---|---|
| 1 | 73.49 | 34.99 | 1.65 |
| 2 | 73.86 | 17.82 | 34.69 |
| 3 | 26.22 | 20.15 | 38.14 |
| 4 | 6.33 | 24.57 | 93.01 |
| 5 | 30.37 | 4.36 | 0.71 |
| 6 | 67.27 | 15.14 | 25.33 |
| 7 | 7.73 | 5.46 | 21.16 |
| 8 | 28.65 | 34.03 | 36.63 |
| 9 | 7.98 | 8.68 | 6.17 |
| 10 | 77.62 | 98.43 | 38.65 |
| 11 | 6.35 | 39.93 | 2.91 |
| 12 | 55.01 | 5.13 | 62.66 |
| 13 | 14.51 | 46.62 | 4.18 |
| 14 | 13.02 | 19.85 | 6.14 |
| 15 | 89.76 | 35.88 | 3.36 |
| 16 | 52.69 | 60.62 | 70.48 |
| 17 | 49.77 | 56.87 | 29.75 |
| 18 | 18.53 | 82.48 | 8.60 |
| 19 | 28.50 | 29.24 | 85.58 |
| 20 | 67.15 | 34.09 | 2.69 |

Table 5: The average rank percentage scores for the baseline rejected samples.

The highest, median and mean ARP values for BRS A were 89.76%, 29.51% and 39.74%, respectively. For BRS B, they were 98.43%, 31.64% and 33.72%, respectively. Finally, for BRS C, they were 93.01%, 23.24% and 28.62%, respectively. Comparing the NUS against each of the three baseline rejected samples using a two-tailed, two-sample t-test assuming unequal variances (TTUV) at a significance level of 0.05, the means were always different to a statistically significant degree, with the mean ARP for the NUS being significantly higher. Table 6 shows the result of each comparison (the mean ARP for the sample is in brackets).

| | **NUS (60.06)** |
|---|---|
| **BRS A (39.74)** | $t(38) = 2.308$, $p = 0.027$ |
| **BRS B (33.72)** | $t(38) = 3.134$, $p = 0.003$ |
| **BRS C (28.62)** | $t(38) = 3.494$, $p = 0.001$ |

Table 6: T-test results of the NUS compared against BRS.

The ARPs of the FENs in the NUS (rightmost column in Table 4) were then combined with the ARPs of the FENs for each BRS (columns 2, 3 and 4 in Table 5). So, there were three columns of 40 ARPs, each one having the ARPs of the NUS in them. Sorted from highest to lowest, the associated FENs in the NUS appeared in the top half 13/20 (65%), 15/20 (75%) and 16/20 (80%) times, respectively. Overall, between 65%-80% of the time (averaging 73.33%), which is well above random chance where only 10/20 or 50% would be expected. In practice, this means that once new FENs are analyzed (and ranked from highest to lowest) based on what the computer has learned from existing ‘liked’ and ‘disliked’ selections, the solver is more likely to be able to find ones they prefer sooner than otherwise, i.e. typically over 70% in just the top half, equivalent to a performance increase of over 40% compared to a typical case of no learning being applied to the new and unseen compositions.

## 5. Discussion and Conclusions

The experimental results presented in the previous section demonstrate that the proposed novel machine learning method is indeed able to learn from existing *single user* preference data, at least in the domain of chess problems. It is worthwhile to point out again that the change from one object to another (see section 3.1), in this case chess positions in the FEN format, has nothing to do with the actual rules of chess but merely the shift from one kind of byte-length character data to another kind (e.g. 'Q' to 'K', '1' to 'b'). This suggests the learning method could also be applied to images or other types of objects with discrete components, perhaps with some modifications or adaptations; but this is beyond the scope of the present article. The way the 'change values' between objects are processed (see section 3.2) implies that the ordering of the objects, i.e. which one comes first, second, third etc. can influence the end result, even though at present this has not been specifically tested for. The learning should therefore be sensitive to how a user's preferences may (and likely will) change over time.

The incremental yet random size selection of the samples to be used (see section 3.3) introduces some variability into the process but this is stabilized by utilizing the *average* rank percentage over three cycles of hundreds, if not thousands of t-tests between the samples for each new FEN under consideration. Even these are flexible aspects of the method that can be modified or adapted to suit particular domains or needs. For example, a fixed but much larger sample size could be used to speed up the process given large databases of user preference data (i.e. liked and disliked material), in which case only one cycle may be necessary, speeding it up even further. On the other hand, doing so may not work as well as the present approach just described; so, it is up to researchers to test for themselves what settings appear to work most reliably in their domains of investigation. Under typical conditions, as was the case here, processing can indeed take many hours.

For instance, 1,000 new and unseen positions processed using a standard desktop computer with user preference databases of about the sizes in our experiment (see section 4) could take around 24 hours to complete or between 1 and 2 full minutes per position. This would still be considered worthwhile, however, since the system could be run in the background while other work is being done. It would save significant amounts of time and energy on the part of the user by not having to go through all 1,000 of those new positions manually but only 500 or so to obtain most of what they would have liked. Given that there are limits to human time and energy, large collections of new and unseen objects are seldom fully and properly assessed anyway. In many cases they are either completely rejected or only a small portion is examined; even then probably less thoroughly as time passes given that most humans tend to tire or bore fairly quickly anyway (McSpadden, 2015).

The automatic sorting of the new objects (assuming more than one is being evaluated) based on their average rank percentages can easily be performed computationally after the process ends so the method is essentially something that can be set and forgotten about until it completes. It does not require much involvement by the user once it is running. As mentioned at the start of this section, the proposed method uses single user preference data and that means it does not rely on external (i.e. other) user preference data or any personal details about the user to be matched with 'other users like them'. Aside from the selection and classification of earlier objects as being 'liked' or 'disliked', noth-ing else from the user is required. User privacy is also better guarded as the machine learning does not require sharing information (it can be done locally on one's own computer using one's own data). It is not yet known what the minimum sizes of the 'liked' and 'disliked' databases should be but since statistical testing is a major component of the proposed method, presumably the larger they are the better. The downside is that the overall process will take longer as these sizes increase. For that reason, the quality of data is important. We suggest that for these databases, only the objects that have actually been seen and properly evaluated by the user be collected. Ambiguous or undecided objects should be avoided and not included.

As time goes by, perhaps some of the earliest objects need not even be included anymore since the user's taste or preferences may have changed significantly enough. A subset of about the sizes used in our experiment (see section 4) or smaller can be randomly selected from both databases if faster processing time is a critical issue. The optimal sample and database sizes in any particular domain, including chess problems, has yet to be determined experimentally, however. This is also worthy of further inquiry. It is noteworthy that, at this point, it remains unclear how the method 'learns' from the 'change values' between objects since they have nothing to do with the actual rules of chess. Our best guess is that there are sufficiently-detectable 'patterns', invisible to humans, that emerge through this process from the databases. Patterns that can be disrupted by new chess problems that somehow do not 'fit' with them, despite a margin of error.

Each pattern would therefore theoretically also be unique to each user and even changes as they change. It is a fascinating idea worthy of further investigation in itself but again, beyond the scope of this article. Another aspect that might be of interest to some is the nature of what is being learned. Even within the domain of chess compositions, is it the aesthetics, the difficulty, the complexity or something else? Our best estimation of the answer is that it captures, to a reasonably good extent, an amalgamation of all that and more, including the 'intangible' and unquantifiable aspects of why individual users like and dislike as they do. The patterns mentioned earlier may be analogous to paths and points in a long and continuing journey that are invisible from the perspective of the one walking (they are making complex yet 'lower-level' decisions of their own from where they stand) yet from a bird's-eye view there is a general but unique pattern to their choices. A kind of swarm intelligence (Schranz et al., 2021), if you will, but involving just a single agent.

Finally, the reader should not lose sight of the fact that this is a novel machine learning approach that still requires more testing (e.g. contrasting learning performance based on many users), even within the domain of chess composition. It may be a challenge to obtain reliable 'liked' and 'disliked' databases of sufficient sizes for single users in most domains, however, and this is a limitation worth pointing out as well. In this article, we have merely presented the general method itself and some experimental results obtained. Regardless of what other researchers may find, we are actually currently using this method within this domain for our own purposes because it does indeed seem to work. It is our hope that we have documented it well enough for other researchers or interested parties to try for themselves using their own user preference data and perhaps even improve upon it in ways we cannot yet imagine at this point.